\documentclass{article}
\usepackage{spconf,amsmath,graphicx}
\usepackage{color, colortbl}
\usepackage{makecell}
\usepackage{multirow}
\title{Multi-Object Tracking using Poisson Multi-Bernoulli Mixture Filtering for Autonomous Vehicles}
\name{Su Pang and Hayder Radha}%\thanks{}}
\address{Michigan State University, East Lansing, MI 48824, USA}
\begin{document}
%\ninept
%
\maketitle
\begin{abstract}
The ability of an autonomous vehicle to perform 3D tracking is essential for safe planing and navigation in cluttered environments. The main challenges for multi-object tracking (MOT) in autonomous driving applications reside in the inherent uncertainties regarding the number of objects, when and where the objects may appear and disappear, and uncertainties regarding objects' states. Random finite set (RFS) based approaches can naturally model these uncertainties accurately and elegantly, and they have been widely used in radar-based tracking applications. In this work, we developed an RFS-based MOT framework for 3D LiDAR data. In partiuclar, we propose a Poisson multi-Bernoulli mixture (PMBM) filter to solve the amodal MOT problem for autonomous driving applications. To the best of our knowledge, this represents a first attempt for employing an RFS-based approach in conjunction with 3D LiDAR data for MOT applications with comprehensive validation using challenging datasets made available by industry leaders.  The superior experimental results of our PMBM tracker on public Waymo and Argoverse datasets clearly illustrate that an RFS-based tracker outperforms many state-of-the-art deep learning-based and Kalman filter-based methods, and consequently, these results indicate a great potential for further exploration of RFS-based frameworks for 3D MOT applications.
\end{abstract}
\section{Introduction}
\label{sec:intro}
\begin{figure}[t]
\begin{center}
   \includegraphics[width=\columnwidth]{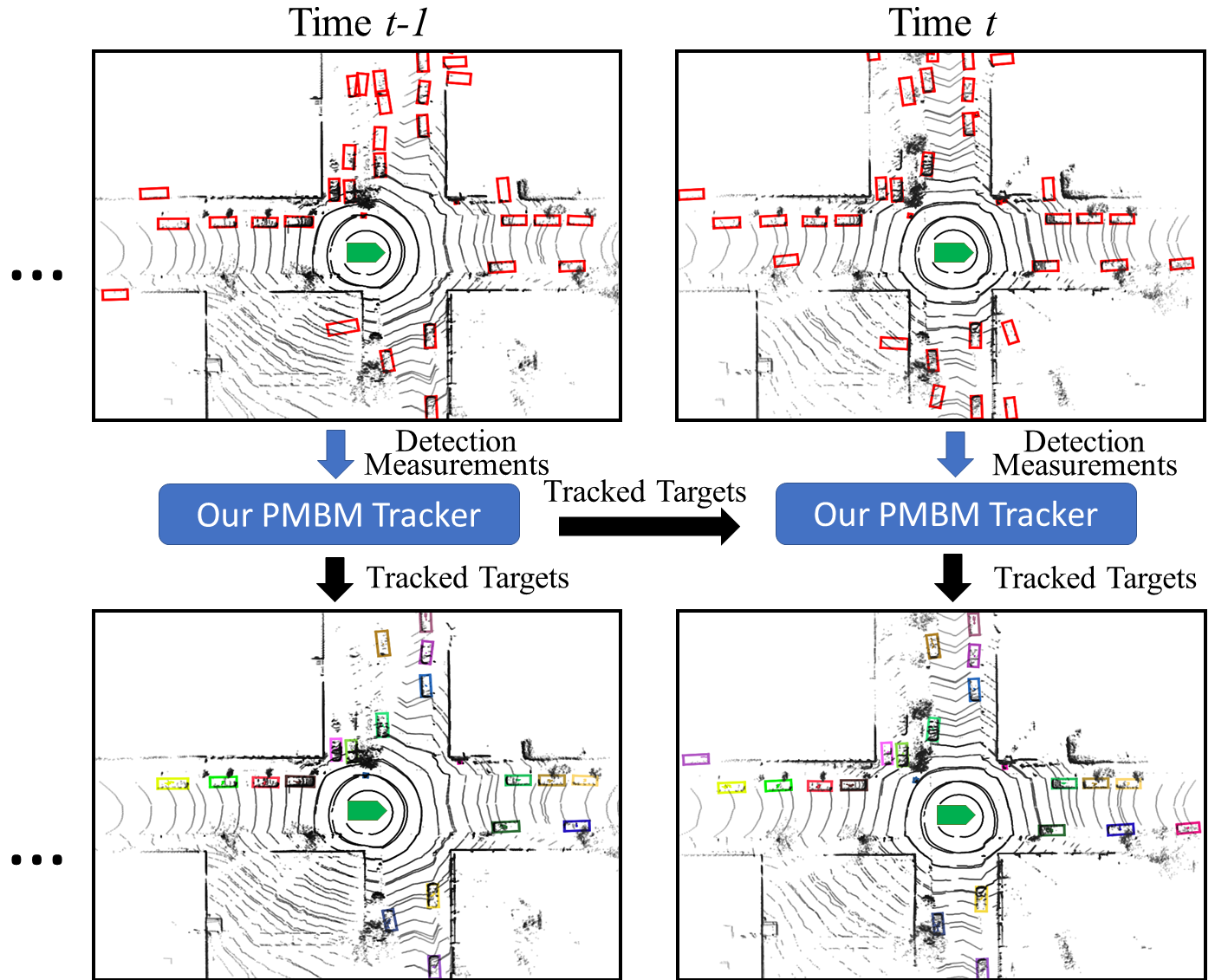}
\end{center}
\vspace{-0.5cm}
   \caption{Overview of the proposed PMBM tracker pipeline. For each frame, many 3D detections are generated by a neural-network-based 3D detector, as the red bounding boxes shown in the top two figures illustrate. Our PMBM tracker could successfully track targets and filter out false positives. In the lower two figures, different bounding box colors correspond to different unique tracked IDs. The green pentagon in the center of each figure represents the ego vehicle pose. All figures are in top-down view. Better viewed in color.}
\label{opening_figure}
\end{figure}
\vspace{-0.3cm}

Multiple object tracking (MOT) is a critical module for enabling an autonomous vehicle achieve robust perception of its environment and consequently achieve safe maneuvering within the environment surrounding the vehicle. The main challenges for MOT in autonomous driving applications are threefold: (1) uncertainty in the number of objects; (2) uncertainty regarding when and where the objects may appear and disappear; (3) uncertainty in objects' states. Traditional filtering based methods, such as Kalman filtering \cite{kalman1960new,weng3d,chiu2020probabilistic}, perform well in state update and estimation but can hardly model the unknown number of objects, and the so-called \textit{birth and death} phenomena of objects. Meanwhile, the emergence of random finite set (RFS)\cite{vo2005sequential,mahler2007statistical,vo2008bayesian} based approaches have opened the door for developing theoretically sound Bayesian frameworks that naturally model all the aforementioned uncertainties accurately and elegantly. 

RFS-based MOT algorithms have been shown to be very effective for radar-based MOT applications \cite{vo2011random,papi2013multi}. In particular, Poisson multi-Bernoulli mixture (PMBM) filtering has shown superior tracking performance and favourable computational cost \cite{xia2017performance} when compared to other RFS-based approaches. Consequently, under this work, we propose a PMBM filter to solve the amodal MOT problem for autonomous driving applications (Fig~\ref{opening_figure}). Applying RFS-based trackers for 3D LiDAR data and/or for 2D/3D amodal detections (bounding boxes) has not been well explored. Existing works in this area either under-perform state-of-the-art trackers or they have been tested using a small dataset that do not reflect broad and truly challenging scenarios ~\cite{kalyan2010random, lee2010tracking, granstrom2017pedestrian}. We believe that RFS-based methods could provide robust and highly-effective solution for these emerging detection modalities.

The contributions of our paper are as follows: (1) We propose a PMBM filter to solve the amodal MOT problem for autonomous driving applications. To the best of our knowledge, this represents a first attempt for employing an RFS-based approach in conjunction with 3D LiDAR data and neural network-based detectors. (2) We demonstrate that our PMBM tracker is low-complexity, and it can run at an average rate of 20 Hz on a standard desktop. (3) We validate and test the performance of our PMBM tracker using two extensive open datasets provided by two industry leaders -- Waymo~\cite{sun2020scalability} and Argoverse~\cite{chang2019argoverse}. These datasets, which have more than 80000 diverse testing frames, clearly demonstrate that our tracker outperforms many state-of-the-art methods under realistic driving conditions. It is worth noting that our PMBM tracker ranks \textbf{No.2} in average MOTA and \textbf{No.1} in vehicle MOTA among all the entries that use the organizer provided detections on the Argoverse dataset.

\begin{figure}[t]
\begin{center}
   \includegraphics[width=0.85\columnwidth]{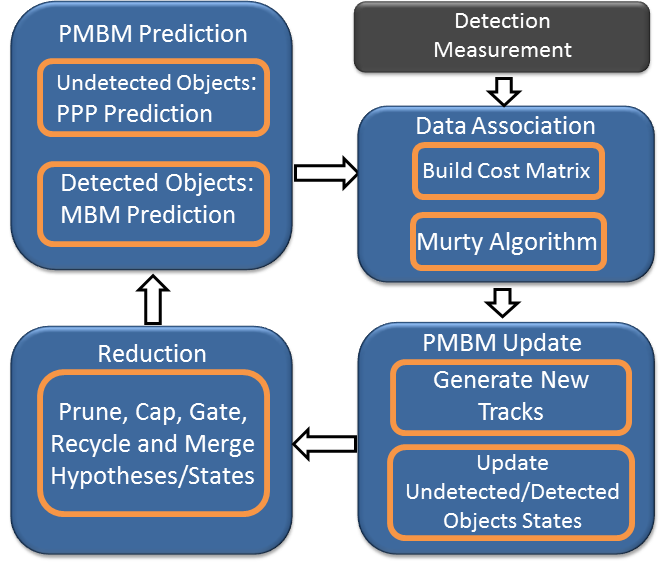}
\end{center}
\vspace{-0.5cm}
   \caption{PMBM tracker system architecture. There are four primary components: (1) PMBM Predictions; (2) Data Association; (3) PMBM Update; and (4) Reduction. PPP represents Poisson point process, and MBM stands for multi-Bernoulli mixture. The detail for each component is provided in Section 2.}

\label{architecture_figure}
\vspace{-0.3cm}
\end{figure}
\vspace{-0.6cm}
\section{Approach}
\vspace{-0.2cm}
\label{sec:format}

The PMBM filter is a relatively new approach to the MOT problem. It is able to track objects based on a Bayesian framework, and it models undetected objects and detected objects using two distinct probability distributions. The high-level system of the proposed PMBM MOT tracker architecture is shown in Fig~\ref{architecture_figure}. As shown in the figure, the tracker consists of four primary components: (1) PMBM Predictions; (2) Data Association; (3) PMBM Update; and (4) Reduction. 

\subsection{Object State}
The object states used in this work is defined as $\mathbf{x}=[x, y, v_x, v_y]$, where $x$ and $y$ represent the 2D location of the object, $v_x$ and $v_y$ are the velocities along $x$ and $y$ directions respectively. The reasons that we define the states in this compressed way are as follows: first, the $z$ value does not change dramatically for consecutive frames; second, the dimension of the objects are already precise from a neural network based detector; therefore, it is not necessary to incorporate all 3D information; third, reducing the state dimension inherently enables the tracking system operate at a lower computational cost for real-time performance. Under this work, our PMBM tracker is designed as a point-based tracker. The center points of the objects, which are initially output from the detector, are tracked using unique tracked IDs, while the bounding box dimensions (height, width, length) are directly extracted from the detection measurements.
\vspace{-0.3cm}
\subsection{Detected and Undetected Objects}
Under the PMBM model, the set of objects $\mathbf{x_t}$ at timestamp $t$ is the union of detected objects $\mathbf{x}^d_t$ and undetected objects $\mathbf{x}^{u}_t$. Detected objects $\mathbf{x}^{d}_t$ are objects that have been detected at least once. Undetected objects $\mathbf{x}^{u}_t$ are objects that are not detected. Note that we are not explicitly tracking the undetected objects, which is impossible under a tracking-by-detection framework. Instead, we have a representation of their possible existences. For example, if we consider an autonomous car under a scenario where a large truck blocks part of the view. It is possible that some objects are located in the occluded area behind the truck, and hence, these objects are inherently undetected.
\vspace{-0.3cm}
\subsection{Data Association Hypotheses}
For each timestamp, there are multiple hypotheses for data association. For our measurement-driven framework, each measurement, either it is a newly detected target, a previously detected target, or a false positive detection. We form different global association hypotheses from possible combinations of the single target hypothesis (STH). Gating is used to reduce the total number of hypotheses and only keeps reasonable ones. Murty's algorithm \cite{murthy1968algorithm}, an extension of the Hungarian algorithm \cite{kuhn1955hungarian} is used to generate $K$ best global hypotheses instead of only one.
\vspace{-0.3cm}
\subsection{PMBM Density}
Under the PMBM model, we use Poisson RFS, also named as Poisson point process (PPP) to represent undetected objects, and multi-Bernoulli mixture (MBM) RFS to represent detected objects. 
The PMBM density is defined as a convolution of a PPP density for undetected objects and a MBM density for detected objects:
\begin{align}
\begin{split}
   \mathcal{PMBM}_t(\mathbf{x}) = \sum_{\mathbf{x^u}\uplus\mathbf{x^d}=\mathbf{x}}\mathcal{P}_t(\mathbf{x^u})\mathcal{MBM}_t(\mathbf{x^d})
\end{split}
\end{align}
where $\mathbf{x}$ represents all the objects in the surveillance area, and where  $\mathbf{x}$ is the disjoint union set of undetected objects $\mathbf{x}^u$ and detected objects $\mathbf{x}^d$. $\mathcal{P}(\cdot)$ and $\mathcal{MBM}(\cdot)$ are the Poisson point process density and multi-Bernoulli mixture density, respectively.
\vspace{-0.2cm}
\subsection{PMBM Prediction}

A crucial aspect of the PMBM filter is its \textit{conjugacy} property, which was proved in \cite{garcia2018poisson}. The notion of conjugacy is quite critical for robust and accurate Bayesian-based MOT. In summary,  the conjugacy of the PMBM filter inplies that if the prior is in a PMBM form, then the distribution after the Bayesian prediction and update steps will be of the same distribution form. Therefore, the prediction stage of a PMBM filter can be written as:
\vspace{-0.2cm}
\begin{align}
\begin{split}
   \mathcal{PMBM}_{t+1|t}(\mathbf{x}_{t+1}) = \int p(\mathbf{x}_{t+1}|\mathbf{x}_t)\mathcal{PMBM}_{t|t}(\mathbf{x}_t)\delta\mathbf{x}_t
\end{split}
\end{align}
where $p(\mathbf{x}_{t+1}|\mathbf{x}_t)$ represents the transition density. Constant velocity model is used as the motion model in this work for simplicity.
Under the PMBM filter, undetected and detected objects can be predicted independently. We define $P_s$ as the probability of survival, which models the probability that an object survives from one time step to the next. For undetected objects, the predicted parameters consist of predicted parameters from the previous timestamp and PPP birth parameters. The weight of each undetected object is scaled by $P_s$ in the prediction step. For detected objects, which are modeled as multi-Bernoulli mixture RFSs, each multi-Bernoulli (MB) process can also be predicted independently of the other MB processes. The probability of existence for each MB-modeled object is decreased by a factor $P_s$ in order to account for the higher uncertainty of existence within the Prediction stage.

\vspace{-0.2cm}
\subsection{PMBM Update}

Furthermore, by adding information from the measurement model $p(\mathbf{z}_{t}|\mathbf{x}_t)$, the PMBM density can be updated with:
%\begin{align}
\vspace{-0.2cm}
\begin{multline}
   \mathcal{PMBM}_{t+1|t+1}(\mathbf{x}_{t+1}) = \\ \frac{p(\mathbf{z}_{t+1}|\mathbf{x}_{t+1}) \mathcal{PMBM}_{t+1|t}(\mathbf{x}_{t+1})}{\int p(\mathbf{x}_{t+1}|\mathbf{x}^\prime_{t+1})\mathcal{PMBM}_{t+1|t}(\mathbf{x}^\prime_{t+1})\delta\mathbf{x}^\prime_{t}}
\end{multline}
%\end{align}
In the update step, the undetected objects that do not have any measurement associated with them remain undetected. The Bayesian update will thus not change the states or variances of the Poisson distributions since no new information is added. Here we define $P_d$ as the probability of detection, which models an object ought to be detected with that probability. For undetected objects without measurement associated, the weight is thus decreased with a factor $(1-P_d)$ as to account for the decreased probability of existing. For detected object, the predicted state is updated by weighting in the information contained in the measurement.

There are two different types of updates for detected objects: the objects being detected for the first time and the detected objects from the previous timestamp. Our tracker is a measurement-driven framework: an object must be connected to a measurement in order to be classified as detected for the first time. All the undetected PPP intensity components and corresponding gated measurements are considered to generate the fused distribution. Note that the detections provided from a neural network always have confident scores attached to them. This confident score is an invaluable indicator of the object probability of existence. So unlike a standard PMBM filter, we incorporate the detection confident score into the update step of objects detected for the first time. We get a new Bernoulli process for each first-time detected object. As for detected objects from the previous timestamp, if there are measurements associated with them, then for each hypothesis, a standard Kalman filter is used to update the state vector, the updated probability of existence is set to 1 because one can not associate a measurement to an object that does not exist; if there is no measurement associated with an object, which was detected from a previous frame, then we maintain the object predicted state unchanged. Furthermore, we decrease the probability of existence and weight with $(1-P_d)$. Note that $P_d$ here is related to the associated detection confident score in the past frames. Here, unlike other standard Kalman filter based trackers, the survival time of detected objects without measurement varies based on the tracking status from the previous time period. 

\definecolor{LightCyan}{rgb}{0.88,1,1}
\begin{table*}[t]
%\resizebox{\textwidth}{!}{
\setlength{\tabcolsep}{1.0pt}
\renewcommand{\arraystretch}{0.8} 
\begin{center}

\begin{tabular}{|c|c|c|c|c|c|c|c|c|}
\hline
{  Method} & {  Split} & {  Class} & {  MOTA (\%) $\uparrow$} & {  MOTP $\downarrow$} & {  \#False Possitive $\downarrow$} & {  \#Misses $\downarrow$} & {  \#IDS $\downarrow$} &  {  \#FRAG $\downarrow$}\\ 
\hline
{  Argoverse Baseline~\cite{chang2019argoverse} }& {  validation } & {  Vehicle} & {  63.15} & {  0.37} & {  17385} & {  19811} & {  122} & {  323}\\
\hline
\rowcolor{LightCyan}
{  Our PMBM Tracker } & {   validation} & {  Vehicle} & {  \textbf{68.70}} & {  0.375} & {  10644} & {  20782} & {  271} & {  1116}\\
\hline
{  Argoverse Baseline~\cite{chang2019argoverse}} & {  validation} & {  pedestrian} & {  41.70} & {  0.257} & {  3783} & {  14411} & {  113} & {  123}\\
\hline
\rowcolor{LightCyan}
Our PMBM Tracker & validation & pedestrian & \textbf{45.50} & 0.249 & 4935 & 11653 & 201 & 903  \\
\hline
Argoverse Baseline~\cite{chang2019argoverse} & test & Vehicle & 65.90 & 0.34 & 15693 & 23594 & 200 & 393\\
\hline
\rowcolor{LightCyan}
Our PMBM Tracker & test & Vehicle & \textbf{71.67} & 0.34 & 8278 & 24165 & 362 & 1217\\
\hline
Argoverse Baseline~\cite{chang2019argoverse} & test & pedestrian & 48.31 & 0.37 & 4933 & 25780 & 424 & 387\\
\hline
\rowcolor{LightCyan}
Our PMBM Tracker & test & pedestrian & \textbf{48.56} & 0.4 & 5924 & 24278 & 783 & 1750  \\
\hline

\end{tabular}
\end{center}
\vspace{-0.5cm}
\caption{Quantitative comparison of 3D MOT evaluation results on Argoverse dataset}
\label{Argo_table}
\end{table*}

\begin{table*}
%\resizebox{\textwidth}{!}{%
\setlength{\tabcolsep}{1.0pt}
\renewcommand{\arraystretch}{0.8} 
\begin{center}

\begin{tabular}{|c|c|c|c|c|c|c|c|}
\hline
{ Method} & {  Split} & {  Class} & {  MOTA (Primary) (\%) $\uparrow$} & {  MOTP $\downarrow$} & {  False Possitive (\%) $\downarrow$} & {  Misses (\%) $\downarrow$}  \\

\hline
{  Waymo Baseline}~\cite{sun2020scalability} & {  test} & All & 25.92 & 0.263 & 13.98 & 64.55    \\
\hline
Argoverse Baseline~\cite{chang2019argoverse} & test & All & 29.14 & 0.270 & 17.14 & 53.47    \\
\hline
Probabilistic KF~\cite{chiu2020probabilistic} & test & All & 36.57 & 0.270 & 8.32 & 54.02    \\
\hline
\rowcolor{LightCyan}
Our PMBM Tracker & test & All & \textbf{38.51} & 0.270 & 7.74 & 52.86 \\
\hline

\end{tabular}
\end{center}
\vspace{-0.5cm}
\caption{Quantitative comparison of 3D MOT evaluation results on Waymo dataset}
\label{Waymo_table}
\vspace{-0.4cm}
\end{table*}

\vspace{-0.3cm}

\subsection{Reduction}
\vspace{-0.2cm}
The assignment problem in MOT will theoretically become NP-hard\cite{emami2018machine}, and hence, the reduction of the number of hypotheses is necessary for decreasing the computational complexity and maintain real-time performance. Five reduction techniques are used in this work: pruning, capping, gating, recycling and merging. \textit{Pruning} is used to remove objects and global hypotheses with low weights. \textit{Capping} is used to set an upper bound for the number of global hypotheses and detected objects. \textit{Gating} refers to limiting the search distance for data association, and where Mahalanobis distance is used here instead of Euclidean distance. \textit{Recycling} is applied to detected objects with lower probability of existence. In that context, instead of discarding these objects, we recycle them by moving them from detected object set to undetected object set. There may be non-unique global hypotheses, and hence, \textit{merging}  would merge these identical global hypotheses into one.

%\subsection{HD Maps}
%Argoverse dataset provides HD maps with binary drivable area and region of interest (ROI) labels at one meter grid resolution. The region of interest is five meters beyond the drivable area. According to the dataset description, all tracking annotations are within the ROI. Therefore, we use the HD map ROI binary grid label as a mask to filter out tracks that are not within the ROI.

%\vfill
%\pagebreak
\vspace{-0.5cm}
\section{Experiment}
\vspace{-0.2cm}
\subsection{Settings}
\vspace{-0.1cm}
\noindent
{\bf Dataset.}~We evaluate our method on two popular open dataset provided by two industry leaders: Waymo \cite{sun2020scalability} and Argoverse~\cite{chang2019argoverse}. Waymo 3D tracking dataset contains 800 training segments, 202 validation segments and 150 testing segments. Each segment includes $\sim$ 200 frames covering 20 seconds. There are $\sim$ 40400 frames in the validation set and $\sim$ 30000 frames in the testing set. Three classes are evaluated: vehicle, pedestrian and cyclist.

Argoverse 3D tracking dataset consists of 113 total number of segments (scenes) and 15-30 seconds for each segment. The data is divided into three sets: 65 segments for training ($\sim$ 13000 frames), 24 segments for validation ($\sim$ 5000 frames) and 24 segments for test ($\sim$ 4200 frames). Since our method is not a learning based method, we don't need to use the training set; but validation set is used for tuning the parameters. Class vehicle and class pedestrian are evaluated. The field of view of these two datasets are both 360 degrees.

Note that for both Waymo and Argoverse datasets, ground truth labels are only available for training and validation sets. For fairness of evaluation of testing samples, one needs to submit the tracking results to the relevant server, and the evaluation results are subsequently published on the corresponding leaderboard.

\noindent
{\bf Detections.}~We use precomputed 3D object detections provided by the dataset organizer. 

\noindent
{\bf Evaluation Metric.}~Standard evaluation metrics \cite{bernardin2008evaluating} for MOT are used. The details of the metrics for each dataset can be found in \cite{sun2020scalability,chang2019argoverse}.

\vspace{-0.5cm}

\subsection{Experimental Results}
The results for Argovers dataset and Waymo dataset are shown in Table~\ref{Argo_table} and Table~\ref{Waymo_table}. As shown in these two tables, our method outperforms other state-of-the-art tracker significantly.
At the time of submission (May 2020), among all the entries that use organizer provided detections, our PMBM tracker ranked \textbf{No.3} in averaged ranking (primary metric), \textbf{No.2} in averaged MOTA for all classes, \textbf{No.1} in Vehicle MOTA, \textbf{No.5} in Pedestrian MOTA within the Argoverse 3D Tracking Leaderboard\footnote{https://evalai.cloudcv.org/web/challenges/challenge-page/453/leaderboard/1278\#leaderboardrank-1}, and \textbf{No.5} on the Waymo 3D tracking leaderboard~\footnote{https://waymo.com/open/challenges/3d-tracking/}. For tracking-by-detection approaches, the impact of the quality of input detections is inherently of a paramount importance. But our performance is still very competitive compared to other entries that use better self-generated detections. %we ranked \textbf{No.4} in averaged ranking (primary metric), \textbf{No.3} in averaged MOTA for all classes, \textbf{No.2} based on Vehicle MOTA, within the Argoverse 3D Tracking Leaderboard and \textbf{No.8} on Waymo 3D tracking leaderboard.
%By the time of submission (June 10th 2020), our PMBM tracker ranked \textbf{No.3} based on Vehicle MOTA, and \textbf{No.4} in averaged MOTA for Vehicle and Pedestrian on Argoverse 3D tracking leaderboard. Our method ranks \textbf{No.8} on the Waymo 3D object tracking leaderboard.

It is worth noting that our tracker has superior performance for the vehicle case compared to pedestrian tracking. We believe that this is attributed to the fact that the distance between pedestrians are much smaller than distances among vehicles. Thus, since our tracker is a point-based method, it is unable to process object dimension, and consequently, close distances between hypotheses could result in merging such hypotheses that should be separated. Subsequently, this could lead to erroneous data associations. Therefore, as shown in Table~\ref{Argo_table}, for class pedestrian, our tracker has higher false positives, ID switches and fragmentations, which are due to low quality data association. Meanwhile, for vehicles, all the vehicle detection measurements are further separated due to the fact that two vehicles cannot be overlaid with each other when compared to the pedestrian case. %In particular, the distance between two vehicle detection center points is usually larger than 1.5 meters. This is due to the fact that two vehicles cannot be overlaid with each other. 

%\subsection{Impact of input detections}
%For tracking-by-detection approaches, the impact of the quality of input detections that are provided by the underlying detector is inherently of a paramount importance. For example, based on our observations and experience in the Waymo 3D tracking competition, the quality of the detections made significant impact on the ranking of the different tracking approaches. In that context, it is worth noting that our approach is among the top three on the leaderboard when considering all the entries that use the organizer provided baseline detections. Meanwhile, one can find some top submissions (more than 10\% better compared to other top submissions) on the Waymo 3D tracking leaderboard using standard Kalman filter. These top submissions employ their own detections, which are significantly more accurate when compared to the organizer provided baseline detections. According to the official Waymo 3D detection leaderboard, the detections of the top performers, who employ their own detectors, are 20-35\% better in average precision compared to the organizer-provided baseline detections. The main challenge of designing a robust tracker is to handle the distortions and errors induced by the input detections. In conclusion, we believe it will be more insightful for evaluating the performance of a tracker under such competition is to employ a consistent detection mechanism.
\vspace{-0.5cm}
\section{Conclusion}
\vspace{-0.3cm}
In this paper, we propose a PMBM filter to solve the 3D amodal MOT problem with 3D LiDAR data for autonomous driving applications. Our framework can naturally model the uncertainties in MOT problem. This represents a first attempt for employing an RFS-based approach in conjunction with 3D LiDAR data, neural network-based detectors and with comprehensive testing in large-scale datasets. The experimental results on Waymo and Argoverse datasets demonstrate that our approach outperforms previous state-of-the-art methods by a large margin. Finally, we hope that our results motivate future research on RFS-based trackers for self-driving applications.

\textbf{Acknowledgement:} 
This work has been supported in part by the Semiconductor Research Corporation (SRC) and by Amazon Robotics under the Amazon Research Award (ARA) program.

\vfill\pagebreak
%\section{REFERENCES}
\label{sec:refs}

\bibliographystyle{IEEEbib}
\bibliography{strings,refs}

\end{document}